\documentclass{article}
\usepackage{graphicx} 
\usepackage{latexsym}
\usepackage{amsmath,amsfonts, amssymb}
\usepackage{mathtools}
\usepackage{amsthm}
\usepackage{algorithm,algorithmic}
\usepackage[colorlinks=true,breaklinks=true,bookmarks=true,urlcolor=blue,
     citecolor=blue,linkcolor=blue,bookmarksopen=false,draft=false]{hyperref}
\usepackage{fullpage}

\newtheorem{theorem}{Theorem}

\newcommand{\n}{\ensuremath{\mathbf n}}
\newcommand{\h}{\ensuremath{\mathbf h}}
\newcommand{\argmin}{\mathop{\rm argmin}}
\newenvironment{myproof}{\proof{Proof.}}{ \endproof}

\title{An Efficient Interior-Point Method for Online Convex Optimization}

\author{Elad Hazan \thanks{Princeton Unviersity and Google DeepMind} \and  Nimrod Megiddo \thanks{IBM Almaden Research Center}}

\date{}

\newtheorem{Def}{Definition}
\newtheorem{Prop}[theorem]{Proposition}
\newtheorem{Cor}[theorem]{Corollary}

\def\mAth{\mathsurround=0pt}
\def\eqalign#1{\,\vcenter{\openup1\jot \mAth
  \ialign{\strut\hfil$\displaystyle{##}$&$\displaystyle{{}##}$\hfil
    \crcr#1\crcr}}\,}

\newcommand{\half}{\mbox{$\frac{1}{2}$}}
\renewcommand{\b}{\ensuremath{\mathbf b}}

\newcommand{\K}{\ensuremath{\mathcal K}}

\newcommand{\reals}{\ensuremath{\mathbb R}}

\newcommand{\ignore}[1]{}

\def\reals{{\mathbb R}}

\renewcommand{\u}{\ensuremath{\mathbf u}}
\renewcommand{\v}{\ensuremath{\mathbf v}}
\newcommand{\interior}{\mathop{\rm int}}

\newcommand{\g}{\ensuremath{\mathbf g}}
\newcommand{\x}{\ensuremath{\mathbf x}}
\newcommand{\y}{\ensuremath{\mathbf y}}
\newcommand{\z}{\ensuremath{\mathbf z}}

\newcommand{\bM}{\ensuremath{\mathcal M}}
\newcommand{\R}{\ensuremath{\mathcal R}}

\def\bB{{B}}

\def\bx{\mathbf{x}}

\def\bA{{A}}
\def\bI{\mathbf{I}}

\def\bI{\mathbf{I}}

\def\bM{\mathbf{M}}

\renewcommand{\R}{\beta}
\renewcommand{\bA}{\mathbf{A}}
\renewcommand{\bB}{\mathbf{B}}
\renewcommand{\bM}{\mathbf{H}}

\begin{document}

\maketitle

\begin{center}
{\small \bf This paper is a journal submission from 2010. \\ Some of the referenced results have since been improved.  } 
\end{center}
\begin{abstract}
A new algorithm for regret minimization in online convex optimization is described. The regret of the algorithm after $T$ time periods is $O(\sqrt{T \log T})$ - which is the minimum possible up to a logarithmic term.
In addition, the new algorithm is adaptive, in the sense that the regret bounds hold not only for the time periods $1,\ldots,T$ but also for every sub-interval $s,s+1,\ldots,t$.
The running time of the algorithm matches that of newly introduced interior point algorithms for regret minimization: in $n$-dimensional space, during each iteration the new algorithm essentially solves a system of linear equations of order $n$, rather than solving some constrained convex optimization problem in
$n$ dimensions and possibly many constraints.

\end{abstract}

\section{Introduction.}

In this paper we present an interior point algorithm for the general online learning framework of online convex optimization with computational and prediction advantages compared to the state of the art. 
The setting of online convex optimization (OCO) is a widely considered framework for online learning capable of modeling a host of prediction problems such as portfolio selection, online routing, prediction from expert advice and more \footnote{we refer the reader to the book and survey \cite{CesaBianchiLugosi06book,Hsurvey10} for more details.}. 

In online convex optimization a decision maker makes a sequence of choices, which are represented as points in a convex set in Euclidean space. After revealing her decision, the  adversary presents the decision maker with a cost function that determines the cost of the decision maker in this particular round. The goal of the decision maker is to minimize regret, or the difference in cost between her aggregate loss and the loss of the best fixed decision in hindsight.

\begin{figure*}[!h]
\includegraphics[width=2.5in]{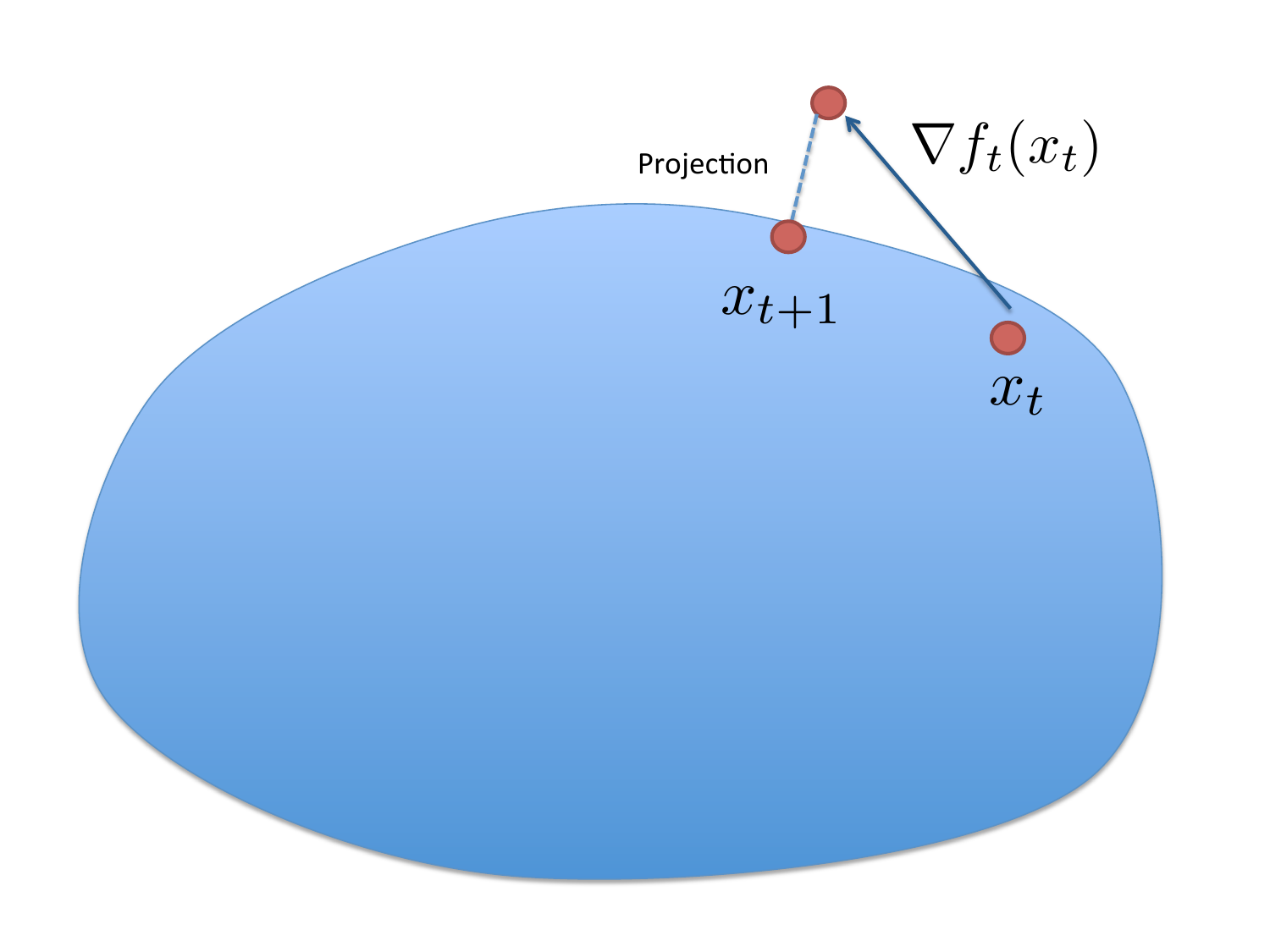} 
\includegraphics[width=2.5in]{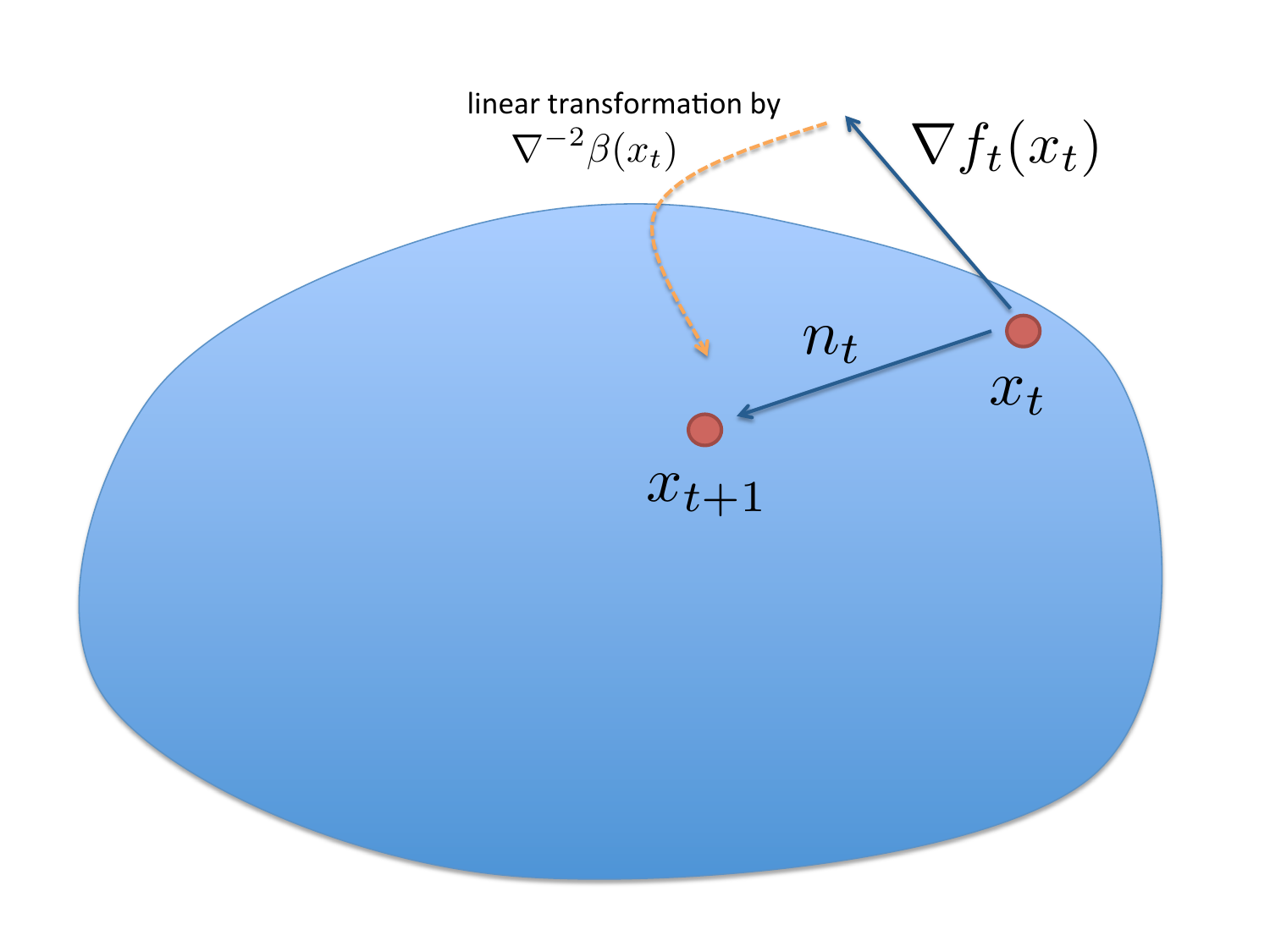} 
\caption{Gradient-based vs. new Interior point algorithm}
\label{fig:algs}
\end{figure*}

The literature of online learning extensively covers polynomial-time algorithms for OCO. Essentially all known algorithms for OCO are variants of the online gradient descent algorithm \cite{Zink} and its generalizations to mirror-decent \cite{Beck2003167,shalev-primal-dual,DBLP:journals/ml/HazanK10} with various Bregman regularizations. The structure of such algorithms is similar: a gradient step is taken with respect to the previous cost function followed by a projection to the convex decision set, as depicted in figure \ref{fig:algs}.

The computation of projections into convex sets can be carried out in polynomial time under reasonable assumptions \footnote{i.e. under assumption that the underlying convex decision set is equipped with a membership oracle, see \cite{Hsurvey10} for more details.}. However, in many scenarios the computation of projections are computationally prohibitive and constitute the most time consuming operation in OCO algorithms. In this paper we present an interior point algorithm for online convex optimization that never leaves the underlying convex set without sacrificing the near-optimal regret guarantees of existing gradient-based methods. 

Our interior point algorithm, depicted in Figure \ref{fig:algs}, is also an iterative algorithm. Every iteration it moves in a direction based on the gradient of the previous cost function. However, instead of moving exactly in the direction of the gradient, which may take us outside of the decision set, the new algorithm first {\it linearly-transforms the gradient according to a self-concordant barrier function of the decision set}. The resulting direction and new point are guaranteed to be inside the decision set. 
Thus, the need for projections is completely removed, and replaced by a linear transformation which is much more efficient to compute.

\subsection{Formal problem statement}
\subsection*{Online convex optimization.}
In the {\em Online Convex Optimization problem} \cite{Zink}, an adversary
picks a sequence of convex functions $f_t: \K \rightarrow \reals$, $t=1,2,\ldots,T$,
where $\K\subset \reals^n$ is convex, compact and bounded. 
At stage $t$, the player has to pick an $\x_t\in \K$ without knowing the function $f_t$.
The player then incurs a cost of $f_t(\x_t)$.
The setting in the paper is that after choosing $\x_t$, the player is informed of the entire
function $f_t$ over $\K$.
The total cost to the player is $\sum_{t=1}^T f_t(\x_t)$. Online Convex Optimization encompasses, for example, experts algorithms
\cite{WarmuthLittlestone89} with arbitrary convex loss function, 
the problem of universal portfolio optimization \cite{cover}, online routing \cite{DBLP:journals/jcss/AwerbuchK08} and more. The reader is referred to \cite{Hsurvey10} for a survey of recent applications in machine learning.

\subsection*{Regret minimization.} 
Suppose the minimum cost over all possible single choices is attained at some
$\x^* =\argmin_{\x\in\K} \sum_{t=1}^T f_t(\x)$.
In this case the {\em regret} resulting from the choices $(\x_1;f_1,\ldots,\x_T;f_T)$ is defined as
\[ R(\x^*) =R(\x_1;f_1,\ldots,\x_T;f_T)~ \equiv~ \mbox{$\sum_{t=1}^T$} \left[f_t(\x_t) - f_t(\x^*)\right]~.\]
The problem of regret minimization calls for choosing the points $\x_1,\ldots,\x_T$ so as to minimize $R$, subject to the
condition that, when $\x_{t+1}$ is chosen, only $\x_1,f_1,\ldots,\x_{t},f_{t}$ are known.
It is known that, in the worst case, the minimum possible regret is $\Omega(\sqrt T)$ (see e.g. \cite{CesaBianchiLugosi06book} or \cite{DBLP:journals/ml/HazanAK07} footnote on page 7.).

\subsection*{Computational efficiency.}
Until recently, known algorithms that attain minimum possible regret
require that in each stage the algorithm solve some constrained convex optimization problem over $\K$,
which can be prohibitive in some practical applications.
In particular, if $\K$ is a convex polyhedron, the best known worst-case bound\footnote{See, for example, \cite{Ye91}.} on the number of
iterations of an optimization algorithm is $O(\sqrt n \, L)$, where $L$ is the number of bits in the description of $\K$,
and each iteration requires solving a linear system of order $n$.

Recently, \cite{AHR08} introduce a novel interior point method which is much more efficient. Building upon their work, we propose a new method for constructing an almost-minimum-regret algorithm, which requires in each stage only solving
a system of linear equations of order $n$, rather than solving an optimization problem over $\K$.
Thus, this method improves the running time at least by a factor
$\sqrt n$, and much more than that when $\K$ is more complicated, for example, a convex polyhedron with many facets.
In contrast to the algorithm of \cite{AHR08}, our algorithm is provably ``adaptive" in the sense that its regret is almost the minimum possible not only over the
stages $1,\ldots,T$ but also over every sub-interval of stages $s,s+1,\ldots,t$. In addition, it does not require an initial computation of the analytic center of the convex set.

\subsection{Previous Algorithms}
There are numerous algorithms for Online Convex Optimization, some of which attain the minimum possible regret of $O(\sqrt{T})$.
Most of these algorithms can be classified into the following two classes:
({\it i}) {\em link-function} algorithms, developed in \cite{GroveLS01, KivinenW01},
and
({\it ii}) {\em regularized follow-the-leader} algorithms (see \cite{Hsurvey10} for bibliography).

\subsection*{Follow-The-Regularized-Leader algorithms.}
The intuitive ``Follow-The-Leader'' \\ (FTL) algorithm picks for $\x_{t+1}$ a minimizer
of the function $F_{t}(\x)\equiv\sum_{s=1}^{t} f_s(\x)$ over $\K$.
It is folklore that the regret of FTL is not optimal \footnote{This can be seen by taking $\K = [-1,1]$ and $f_t$ to alternate between $f_t(x) = x$ or $f_t = -x$. FTL will alternate between $-1$ and $1$, always suffering maximal loss. In contrast - staying with either point in the long run attains half the loss of FTL.  See also \cite{Hsurvey10} page 6.}.
This fact suggested the more general ``Follow-The-Regularized-Leader'' (FTRL) algorithm that picks
$\x_{t+1}$ as a minimizer of the function $F_{t}(\x) + \rho(\x)$ over $\K$,
where $\rho(\x)$ is a certain function that serves as a ``regularizer''.
Different variants of the method correspond to different choices of $\rho(\x)$.
The FTRL approach led to the resolution of some prediction problems, notably the resolution of
the value of bandit information in \cite{AHR08}.
 One advantage of the FTRL approach is its relatively intuitive analysis due to Kalai and Vempala \cite{KV-FTL}.
On the negative side, FTRL algorithms are known to be ``non-adaptive", in the sense that the
regret over a general sub-interval $s,s+1,\ldots,t$ may be linear in $t-s$ rather
than $O(\sqrt{t-s})$ \cite{DBLP:conf/icml/HazanS09}. Furthermore, the running time of the algorithm in each stage may not
be practical because the algorithm has to solve
some optimization problem over $\K$.

\subsection*{The ``link function" methodology.}

In contrast to the intuitive FTRL methodology, which relies on the entire past of history of the play,
link-function algorithms, as developed in \cite{GroveLS01,KivinenW01}, use less information and proceed ``incrementally". They are also known as ``primal-dual" or ``online mirrored descent" algorithms, in analogy to their offline counterparts - mirrored descent algorithms for convex optimization \cite{NemYubook}.
Perhaps the easiest algorithm to describe is Zinkevich's \cite{Zink} online gradient descent, which picks $\x_{t+1}$ to be
the orthogonal projection of the point $ \y_{t+1}\equiv \x_{t} - \eta \nabla f_{t}(\x_{t})$ into $\K$.
Of course, $\x_{t+1}$ is the point in $\K$ nearest $\y_{t+1}$, hence its computation can be costly, for example, if $\K$ has many facets.
On the other hand, link-function algorithms are adaptive (in the sense explained above)
and are usually more efficiently computable than FTRL algorithms in case projections turn out to be easy.

\subsection*{Merging the two approaches.}
The main idea in the new algorithm is to follow the incremental-update approach, but make sure it
never requires projections from the exterior of $\K$ into $\K$ (hence the name ``interior point").
This is accomplished by moving from $\x_{t}$ to $\x_{t+1}$ in a direction that is obtained from the
gradient of $\nabla f_{t}(\x_{t})$ by a linear transformation (like in Newton's method),
which depends on $\K$ and $\x_{t}$.
The assumption is that $\K$ is specified by means of a self-concordant barrier function (see below).
This idea was introduced to learning theory by Abernethy, Hazan and Rakhlin \cite{AHR08}, who used these barriers as regularizers.
Our technique can be interpreted as {\em using the barrier function as a link function rather than a regularizer}.

\subsection{The new contribution compared to previous work}

The contribution of this paper is the design and analysis of a new algorithm for online
convex optimization. The regret of the algorithm is almost the minimum possible.
It is adaptive and requires only solving one system of linear equations of order $n$ per stage.
In comparison to previous work, most other minimum-regret algorithms require, in the worst case, solving a complete optimization problem each iteration.
The one exception which matches this running time is \cite[Algorithm 2]{AHR08}.
It has minimum regret and its computation complexity is close to that of our algorithm.
However, \cite[Algorithm 2]{AHR08} is not adaptive. In addition, it requires computing of the so-called analytic center of $\K$ for the starting point $\x_1$, which requires solving a nontrivial optimization problem, and as originally defined works only for linear cost functions (although this is correctable by taking the gradient approximation of the loss functions). 
We prove that an efficient interior point algorithm can be derived via the ``link function" methodology together with tools from interior point methods. Previously, such results were proven only from the ``follow the leader" methodology. 

\section{Preliminaries.} 
In the following we denote for two points $\x,\y \in \reals^n$ by $[\x,\y]$ the line segment between them, or formally:
$$ [\x,\y] = \{ \z \in \reals^n \ | \ \z = \alpha \x + (1-\alpha) \y \ , \ \alpha \in [0,1] \} $$
\subsection*{Self-concordant barrier.}
We assume that $\K$ is given by means of a {\em barrier function}
$ \R:\interior\K\rightarrow\reals$,
i.e., for every sequence $\{\x^k\}_{k=1}^\infty \subset \interior\K$ that converges to
the boundary of $\K$, the sequence $\{\R(\x^k)\}$ tends to infinity.
We further assume that for some $\vartheta > 0$,
$\R(\x)$ is a {\em $\vartheta$-self-concordant barrier} \cite{Nemi},
i.e., it is thrice differentiable and for every $\x\in\interior\K$ and every $\h\in\reals^n$,
the function $\tilde f(t)=\tilde f(t;\x,\h)\equiv \R(\x+t\h)$ satisfies
\ignore{
\begin{enumerate}
\item [({\it i}\/)]
$|\tilde f'''(0)| \le 2 [\tilde f''(0)]^{3/2}$ (i.e., $\tilde f$ is a self-concordant function), and also
\item [({\it ii}\/)]
$[\tilde f'(0)]^2 \le \vartheta \cdot \tilde f''(0)$.
\end{enumerate}
}
({\it i})
$|\tilde f'''(0)| \le 2 [\tilde f''(0)]^{3/2}$ (i.e., $\tilde f$ is a self-concordant function), and also
({\it ii})
$[\tilde f'(0)]^2 \le \vartheta \cdot \tilde f''(0)$.
It follows that $\R(\x)$ is strictly convex. For example, for $\bA\in\reals^{m\times n}$ and $\b\in\reals^m$, the function
$\R(\x) \equiv - \sum_{i=1}^m \ln [(\bA\x)_i - b_i ]$,
(defined for $\x$ such that $\bA\x > \b$)
is an $m$-self-concordant barrier for the polyhedron $\{\x\in\reals^n~|~\bA\x \ge \b\}$.

\subsection*{The Dikin ellipsoid.}  \label{sec:self_concordant}
For every $\v \in \reals^n$ and $\bA \in \reals^{n\times n}$, denote
$ \|\v\|_\bA \equiv \sqrt{\v^{\top} \bA \v}$, and
for every $\h \in \reals^n$, denote
$ \|\h \|_{\x} \equiv \sqrt{ \h^{\top} [\nabla^2 \R(\x)]  \h }$.
The open \emph{Dikin ellipsoid} of radius $r$ centered at $\x$, denoted by
$W_r(\x)$, is the set of all $ \y=\x+\h\in \K $ such that
 $ \|\h\|_{\x}^2  =   \h^{\top} [\nabla^2 \R(\x)] \h  < r^2 $.

Below we use the following known facts about the Dikin ellipsoid and self-concordant
functions, whose proofs can be found in \cite[Section 2.2, p.~23]{Nemi}.

\begin{Prop}  \label{fact:1}
For every~ $\x\in \K$, $ W_1 (\x) \subset \interior\K$.
\end{Prop}
The next proposition provides ``bounds'' on the Hessian $\nabla^2\R(\x+\h)$ of $\R(\x+\h))$ within Dikin's ellipsoid.
For $\bA,\bB \in \reals^{n\times n}$, the notation
$\bA  \succeq \bB$ means that $\bA - \bB$ is positive
semi-definite.
\begin{Prop}  \label{fact:2}
For every $\h$ such that $\|\h\|_{\x} < 1$,
 \begin{equation}
    \label{eqn:prop_hessian}
    (1-\|\h\|_{\x})^2 \nabla^2 \R (\x)
    ~\preceq ~
    \nabla^2 \R(\x+ \h)
    ~\preceq~
    (1-\|\h\|_{\x})^{-2} \nabla^2 \R(\x)~ .
\end{equation}
\end{Prop}

We denote the diameter of $\K$ by $\Delta$.
\begin{Prop} \label{eigens1}
If
{\rm(}i\,{\rm)} $\R(\x)$  is a barrier function for $\K$, and
{\rm(}ii\,{\rm)} $\R(\x)$ is self-concordant,
then for every $\x\in \interior\K$, all the eigenvalues of $\nabla^2\R(\x)$
are greater than or equal to $\frac{1}{\Delta^2}$.
\end{Prop}
\begin{myproof}
Fix $\x\in\interior\K$ and a {\em unit} vector $\u\in\reals^n$.
Denote
\begin{equation} \label{def:tmax}
 t_{\max} = t_{\max}(\x,\u) \equiv \max\{ t\,|\, \x+t\,\u\in\K\}\le \Delta ~.
 \end{equation}
Define a function $\tilde\beta:[0,t_{\max})\rightarrow\reals$ by
\begin{equation}  \label{def:beta}
\tilde\beta(t) = \tilde\beta(t;\x,\u) \equiv  \R(\x+t\,\u)~.
\end{equation}
Note that
$\tilde\beta'(t)   = \u^{\top}[\nabla\R(\x+t\u)]$
and
$ \tilde\beta''(t)  = \u^{\top}[\nabla^2\R(\x+t\u)]\u $.
Since $\R(\x)$ is self-concordant,
for every $t\in[0,t_{\max})$,
$ \bigg|\left[(\tilde\beta''(t))^{-\half}\right]' \bigg|
 =  \frac{|\tilde\beta'''(t)|}{2[\tilde\beta''(t)]^{3/2}} \le 1$.
 It follows that
 $ t \ge
 \bigg| \int_0^{t} \frac{\tilde\beta'''(s)}{2[\tilde\beta''(s)]^{3/2}}\,ds\bigg|
 = \left|[\tilde\beta''(t)]^{-\half} - [\tilde\beta''(0)]^{-\half} \right|  $.
Now, as $t$ tends to $t_{\max}$, $\tilde\beta(t)$ tends to infinity,
and hence also $\tilde\beta'(t)$  and $\tilde\beta''(t)$ tend to infinity;
thus, $[\tilde\beta''(t)]^{-\half}$ tends to zero.  It follows that
$ [\tilde\beta''(0)]^{-\half} \le t_{\max} \le \Delta $
and hence
$
\u^{\top}\nabla^2\R(\x)\u
= \tilde\beta''(0) \ge \frac{1}{\Delta^2}$.
Since this holds for every unit vector $\u$, the claim follows.
\end{myproof}
\begin{Cor}  \label{Cor:inverseH}
For every $\x\in\interior\K$, all the eignevalues of $[\nabla^2\R(\x)]^{-1}$ are less than or equal to $\Delta^2$.
\end{Cor}

\section{Algorithm and regret bounds}

\subsection{The algorithm and its performance.}
We assume in this section that when the player has to pick the next point $\x_{t+1}$,
the player recalls $\x_t$ and knows $\nabla f_t(\x_t)$ and $\nabla^2\R(\x_t)$.
Interior-point algorithms for optimization typically utilize the Newton direction. 
In the case of minimizing a function of the form
 $F_\mu(\x) \equiv f(\x) - \mu \cdot\R(\x)$,
while at a point $\x$, the newton direction would be
$ \n =-[ \nabla^2 F_\mu(\x)]^{-1} \nabla(F_\mu)(\x)$.
However, for minimum-regret online optimization, it turns out that the following direction is
useful:
\[ \n_t = -  [\nabla^2\R(\x_t)]^{-1}\nabla f_t(\x_t)~,\]
i.e., the gradient factor is determined by the previous objective function $f_t$,
while the Hessian factor is determined by the barrier function $\R$. This is reminiscent to the online gradient descent algorithm of \cite{Zink}, in which only the gradient of the last cost function is used, although we also use the linear transformation defined by the barrier.  The standard Newton direction is not used since our objective in online convex optimization is to minimize regret rather than reach a global optimum. 

Thus, when our algorithm is employed, the player picks
$ \x_{t+1} = \x_t + \eta\,\n_t$,
where $0< \eta < 1$ is a scalar whose value depends on $T$; it tends to zero as $T$ tends to infinity.
Denote
$\g_t = \nabla f_t(\x_t) $
and
$\bM_t = \nabla^2\R(\x_t) $.
Hence,
$\n_t = -\bM_t^{-1}\g_t$.

\begin{algorithm}[H] \caption{Online IP-step \label{alg:main}}
\begin{algorithmic} [1]
\STATE Input: $\vartheta$-self-concordant barrier $\beta$ for $\K \subseteq \reals^n$, learning rate $\eta >0$. 
\STATE Initialize $\x_1$ arbitrarily. 
\FOR{$t=1, 2, \ldots, T$}
\STATE Play $\x_t$ and observe $f_t$.
\STATE Set $\x_{t+1} = \x_{t} + \eta \n_t$.
\ENDFOR
\end{algorithmic}
\end{algorithm}

Our main theorem bounds the regret of Algorithm \ref{alg:main} by $O(\sqrt{T \log T})$, where the $O$ notation hides constants of the setting independent of $T$. The formal statement is the following theorem, which we proceed to  prove it in the rest of this section.

\begin{theorem} \label{thm:main}
Suppose that  $ \eta = \frac{D}{\sqrt{10} G \Delta }\cdot \frac{1}{\sqrt T}$  for $D = \sqrt{\vartheta \ln T  + G\Delta} $, and $\eta G \Delta \leq \frac{1}{4}$. Then for every $\x^*\in\K$, Algorithm \ref{alg:main} attains
\begin{eqnarray*}
 R(\x^*) & \leq 9 \sqrt{\vartheta}  G \Delta \sqrt{T \log T} + 9  (G \Delta)^{3/2}   \sqrt{T} \\ 
& = O( \sqrt{\vartheta}  G\Delta \sqrt{T\log T} )  
\end{eqnarray*}
\end{theorem}

\subsection{Validity.}
We first prove that the algorithm generates only points in $\K$.
\begin{Prop}
For every $t$, if $\x_t\in \interior \K$ and
$ \eta < (\g_t^{\top} \bM_t^{-1}\g_t)^{-\half}$,
then
$ \x_{t+1} \in\interior \K$.
\end{Prop}
\begin{myproof}
In view of Proposition \ref{fact:1}, it suffices to prove that
$\x_{t+1} \in W_1(\x_t)$.
By the definition of the Dikin ellipsoid, it suffices to show that
$  \|\eta \n_t\|_{\bx_t}
   =  \sqrt{ (\eta \n_t)^{\top} [\nabla^2 \R(\x_t)] (\eta \n_t) } < 1 $.
Indeed,
\[
\| \x_{t+1} - \x_t\|_{\bx_t}^2
\equiv
 (\x_{t+1} - \x_t)^\top \bM_t  (\x_{t+1} - \x_t)
=   \eta^2\, \n_t^{\top} \bM_t \n_t
=  \eta^2\, \g_t^{\top}  \bM_t^{-1} \g_t
 < 1 ~.\]
Thus, $\x_{t+1}\in W_1(\x_t)\subset \interior \K$.
\end{myproof}
By Corollary \ref{Cor:inverseH},
\begin{equation} \label{eq:bdgrad}
  \g_t^{\top} \bM_t^{-1}\g_t \le \Delta^2\cdot\|\g_t\|^2 ~.
\end{equation}
Thus, we also have
\begin{Cor} \label{cor:validity}
If
$ \eta <  \frac{1}{\Delta\cdot\|\g_t\|}$,
then
$\x_{t+1} \in\interior \K$.
\end{Cor}
\subsection{A bound on the gradients.} 
As with any first-order method, the smoothness of the objective function, or alternatively an upper bound on the gradient size, plays a role in its convergence rate. The convexity and boundedness properties of the loss functions  also affect regret minimization, and thus we wish to express our bound on the regret with
respect to bounds on the gradients of the functions selected by the adversary.
Thus, we denote
\[ G = \max  \left\{ \| \nabla f_t(\x)\|\,:\,\x\in\K,\ t=1,\ldots,T \right\}  ~.\]
Since the player does not know the function $f_t$ at the time of picking $\x_t$, and
that choice depends on $G$, we simply assume that the adversary is restricted to
choosing only functions $f$ such that $\|\nabla f(\x)\| \le G$ for every $\x\in\K$.

We note that standard techniques, in particular the ``doubling trick", can be used,
without harming our asymptotic regret bounds, to eliminate the requirement
that the algorithm knows an upper bound $G$ a priori. Roughly speaking, this simple technique assumes an a-priory fixed upper bound on $G$. If during the repeated game this bound is exceeded, then the assumed upper bound is doubled. It can be shown that the overall regret forms a geometric series, and this way our results can be extended to the case that $G$ is not known in advance.

\begin{Prop} \label{prop:7eta}
Assuming $\eta G \Delta \leq \frac{1}{4}$, then for every  $t=1,\ldots,T$, and every $\x^* \in \K$ 
\begin{equation}  \label{eq:7eta}
 \eta \,\g_t^{\top} (\x_t - \x^*)
~\leq~  [\nabla \R(\x_{t+1}) - \nabla \R(\x_t)]^{\top} (\x^* - \x_t)
 + 8 G^2 \Delta^2 \cdot \eta^2  
\end{equation}
\end{Prop}
Before diving into the proof, we note that in our algorithm $\eta$ is indeed chosen to satisfy the assumption of this hypothesis. 
\begin{myproof}
Denote $\v = \x_t - \x^*$.
Consider the function $\gamma:\reals^n\rightarrow \reals$ defined by
$\gamma(\y) = [\nabla \R(\y)]^{\top} \v$.
Note that
$\nabla\gamma(\y) = [\nabla^2 \R(\y)] \v$.
By Taylor's theorem, there exists a point $\z\in[\x_t,\x_{t+1}]$ such that
$
 \gamma(\x_{t+1}) - \gamma(\x_t)
=   (\x_{t+1} - \x_{t})^{\top} \nabla \gamma(\z)
=  [-\eta\bM_t^{-1} \g_t]^{\top} [\nabla^2 \R(\z)] \v
=  -\eta \g_t^{\top}  \v - \eta \g_t^{\top} \bM_t^{-1} [\nabla^2 \R(\z) - \bM_t] \v $.
Denote $\bM_{\z} = \nabla^2 \R(\z)$ and $\bB = \bM^{-1}_t \bM_\z - \bI$.
Thus,
\begin{equation} \label{eqn:shalom1}
\eta \g_t^{\top} \v + [\nabla \R(\x_{t+1}) - \nabla \R(\x_{t})]^{\top} \v
   =  -\eta\, \g_t^{\top} \bM_t^{-1} (\bM_{\z} - \bM_t) \v
   =  -\eta  \g_t^{\top} \bB \v ~,
\end{equation}
We proceed to bound the LHS of the above. Notice that:
\begin{eqnarray}
 (\g_t^{\top} \bB \v )^2  & =  (\g_t^{\top} \bB \v )  (\v^{\top} \bB^T \g_t^{\top}  ) = \g_t^{\top} \bB \v \v^{\top} \bB^T \g_t^{\top}  \notag \\
 & \leq \| \g_t \|^2 \| \bB \v \v^{\top} \bB^T  \| & \mbox { $\| B \|$ denotes spectral norm} \notag  \\
 & \leq G^2 \| \bB  (\|\v\|^2 \bI )  \bB^T  \| & \mbox{ $BA B^T \preceq B C B^T$ for $A \preceq C$} \notag \\
 & = G^2 \|\v\|^2  \| \bB  \bB^T  \| \notag  \\
 & \leq 4 G^2 \Delta^2  \| \bB  \bB^T  \| & \mbox{ defn of $\v = \x_t - \x^* $} \label{eqn:shalom3}
 \end{eqnarray}
We proceed to bound $\|\bB \bB^T\|$

Since $\z \in [\x_t,\x_{t+1}]$, $\x_{t+1} \in W_1(\x_t)$,  and $W_1(\x_t)$ is convex, it follows
that $\z \in W_1(\x_t)$.
Thus, by \eqref{eqn:prop_hessian},
$
 (1- \|\z-\x_t\|_{\bM_t})^2\bM_t \preceq \bM_\z \preceq ~(1- \|\z-\x_t\|_{\bM_t})^{-2}\bM_t
$.
Since $\z - \x_t = \alpha \cdot (\x_{t+1} - \x_t)$ for some $\alpha\in[0,1]$, it follows that
$ \|\z-\x_t\|_{\bM_t} = \alpha\|\x_{t+1}-\x_t\|_{\bM_t}$
and hence
$
 (1- \|\x_{t+1}-\x_t\|_{\bM_t})^2\bM_t \preceq \bM_\z \preceq ~(1- \|\x_{t+1}-\x_t\|_{\bM_t})^{-2}\bM_t
$.
Equivalently,
\begin{equation} \label{eq:9a}
 \left(1- \eta\sqrt{\g_t^{\top}\bM^{-1}_t\g_t}\right)^2\bM_t
 \preceq \bM_\z
 \preceq ~\left(1- \eta\sqrt{\g_t^{\top}\bM^{-1}_t\g_t}\right)^{-2}\bM_t~.
\end{equation}
Since by Corollary \ref{Cor:inverseH}
$\sqrt{\g_t^{\top}\bM^{-1}_t\g_t} ~\le~   \|\g_t\|\cdot\Delta  ~\le~ G\cdot\Delta$, and we assume $\eta G \Delta < \frac{1}{4} < 1$,
it follows from (\ref{eq:9a}) that
\begin{equation}  \label{eq:9}
 (1- \eta G\Delta)^2\bM_t
 \preceq \bM_\z
 \preceq ~(1- \eta G\Delta)^{-2}\bM_t~.
\end{equation}

Obviously, if a matrix $\bM$ is positive semi-definite, then so is $\bA^{\top}\bM\bA$ for every matrix $\bA$.
Thus, if $\bM \succeq {\mathbf {M}}$, then for every matrix $\bA$, $\bA^{\top}\bM\bA \succeq \bA^{\top}{\mathbf {M}}\bA$.
Since $\bM_t$ is positive-definite, there exists an $\bA\in\reals^{n\times n}$
such that $\bM_t^{-1} = \bA\bA^{\top}$, so $\bA^{\top}\bM_t\bA =\bI$.
Thus,
\begin{equation} \label{eq:prec3}
(1- \eta G\Delta)^{2}\bI
\preceq \bA^{\top}\bM_\z \bA
\preceq
(1- \eta G\Delta)^{-2}\bI ~.
 \end{equation}
We now bound the eigenvalues of
$\bB
=  \bM^{-1}_t \bM_\z - \bI
=  \bA\bA^{\top} \bM_\z \bA\bA^{-1} - \bI
=  \bA(\bA^{\top} \bM_\z \bA -\bI)\bA^{-1} $.
Recall that a matrix $\bM$ and a matrix ${\mathbf {M}}^{-1}\bM{\mathbf {M}}$ have the same eigenvalues.
Thus,
$\lambda_{\max}(\bB) = \lambda_{\max}(\bA^{\top} \bM_\z \bA -\bI) = \lambda_{\max}(\bA^{\top} \bM_\z \bA) - 1$
and
$\lambda_{\min}(\bB) = \lambda_{\min}(\bA^{\top} \bM_\z \bA) - 1  $.
It follows  from (\ref{eq:prec3}) that all the eigenvalues of
$\bA^{\top} \bM_\z \bA$ are between $(1- \eta G\Delta)^{2} $ and $(1- \eta G\Delta)^{-2}$; hence,
all the eigenvalues of $\bB$ are between
$  \ell \equiv (1- \eta G\Delta)^{2} -1 =  -2 \eta G \Delta+ (\eta G\Delta)^2  \geq -2 \eta G \Delta  $
and
$ u \equiv (1- \eta G\Delta)^{-2}-1
= \frac{2\eta G\Delta - (\eta G\Delta)^2}{1-2\eta G \Delta+ (\eta G \Delta)^2} \le
  \frac{2\eta G \Delta}{1-2\eta G \Delta} $.
Assuming $\eta G\Delta < 1/4$, we have
$ u < 4\eta G \Delta $. 

Thus, 
\begin{eqnarray*}
\|\bB \bB^T\| &\leq \max\{u^2,l^2\} \leq 16 \eta^2 G^2 \Delta^2 
\end{eqnarray*}
Plugging back into \ref{eqn:shalom3} we get 
$$ \left| \g_t^{\top} \bB \v \right| \leq 2 G \Delta \cdot \eta 4 G \Delta \leq \eta 8 G^2 \Delta^2 $$


Plugging the latter into \eqref{eqn:shalom1}, we get
\[
 \eqalign{
\eta \g_t^{\top} \v +& [\nabla \R(\x_{t+1}) - \nabla \R(\x_{t})]^{\top} \v
=  - \eta\g^{\top}\bB\v
\leq 8 \eta^2   G^2  \Delta^2 
}  
\]
\end{myproof}

\subsection{A bound dependent on Bregman divergence.}

For this section, denote by  $\x_1,\ldots,\x_T$  sequence that is generated by Algorithm \ref{alg:main}. Given the functions $f_1,\ldots,f_T$ and the choices $\x_1,\ldots,\x_T$,
recall that for any $\x^*\in \K$, the {\em regret with respect to $\x^*$} is given by
\[ R(\x^*)\equiv \mbox{$\sum_{t=1}^T f_t(\x_t) - \sum_{t=1}^T f_t(\x^*)$} ~.\]

For $\x,\y\in \interior \K$, the {\bf Bregman divergence} $B_\R(\x,\y)$ with respect to the barrier  $\R(\x)$ is given by
\[ B_\R(\x,\y) = \R(\x) - \R(\y) - [\nabla \R(\y)]^{\top} (\x-\y) ~.\]
We note that this divergence is well defined for any convex function, not necessarily a barrier. 

Further, in this section we  assume that we know a bound on the Bregman divergence as follows. Although the player does not know the
optimal point $\x^*$, for the purpose of design and analysis of the algorithm we use a bound $D$ that is guaranteed to satisfy
\[ D \ge \sqrt{B_\R(\x^*,\x_1)} ~\]
The parameter $D$ will be used in the choice of the learning rate $\eta$ in Algorithm \ref{alg:main}, and an explicit choice of $D$ will be used to obtain the regret bound given by Theorem \ref{thm:main}. 

\begin{theorem}
\label{thm:general_norm_past_curvature}
Suppose that  $ \eta = \frac{D}{\sqrt{10} G \Delta }\cdot \frac{1}{\sqrt T}$  and $\eta G \Delta \leq \frac{1}{4}$. Then Algorithm \ref{alg:main} generates valid points in the convex domain and satisfies for every $\x^*\in \interior\K$ for which  $ D \ge \sqrt{B_\R(\x^*,\x_1)} $, 
\[ R(\x^*) \leq 8 D G\Delta \cdot \sqrt T~. \]
\end{theorem}
\begin{myproof}
The proof uses ideas from the proof of \cite[Theorem 4.1]{AHR08} which, in turn, follows
\cite{DBLP:journals/ml/HazanAK07}, with the Bregman divergence
replacing the Euclidean distance as a potential function.

Notice that the assumptions of Corollary \ref{cor:validity} are satisfied by choice of $\eta$. 

Let $\x^* \in \interior \K$ be fixed.
Since the functions $f_t$ are convex,
\begin{equation} \label{conv}
 f_t(\x_t) - f_t(\x^*) \leq [\nabla f_t(\x_t)]^{\top} (\x_t - \x^*) ~ .
\end{equation}
Now,
\begin{equation} \label{3bregman}
 \eqalign{
B_\R(\x,\y) -  B_\R(\x,\z) + B_\R(\y,\z)
=&\  - [\nabla \R(\y)]^{\top} (\x-\y) + [\nabla \R(\z)]^{\top} (\x-\y) \cr
=&\ (\x-\y)^{\top} [\nabla \R(\z) - \nabla \R(\y)] ~.}
\end{equation}
Hence, by (\ref{conv}), (\ref{3bregman}) and Proposition \ref{prop:7eta} (whose assumptions are satisfied by choice of $\eta$)
\begin{equation} \label{1regret}
 \eqalign{
 f_t(\x_t)& - f_t(\x^*)
\leq   [\nabla f_t(\x_t)]^{\top} (\x_t - \x^*)  \cr
\leq &\  \mbox{$\frac{1}{\eta}$} \cdot [\nabla \R(\x_{t+1}) - \nabla \R(\x_{t})]^{\top} (\x^* - \x_t)
+ 8 G^2 \Delta^2  \, \eta \cr
= &\ \mbox{$\frac{1}{\eta}$}\, \left[B_\R(\x^*,\x_{t})-B_\R(\x^*,\x_{t+1}) + B_\R(\x_t,\x_{t+1})\right]
   + 8 G^2 \Delta^2  \, \eta  ~.\cr
     }\end{equation}

Summing (\ref{1regret}) over all stages,
\begin{equation} \label{eq:general1}
\eqalign{
R(\x^*)
\leq &\ \mbox{$\frac{1}{\eta} \left[ B_\R(\x^*,\x_1) - B_\R(\x^*,\x_T) \right]
  + \frac{1}{\eta}  \sum_{t=1}^T  B_\R(\x_t,\x_{t+1})
 + T\,  8 G^2 \Delta^2 \,\eta $} \cr
\leq &\ \mbox{$\frac{D^2}{\eta} + \frac{1}{\eta}\cdot \sum_{t=1}^T  B_\R(\x_t,\x_{t+1})
  + T\cdot  8 G^2 \Delta^2 \, \eta$} ~.
}
\end{equation}
To bound $B_\R(\x_t,\x_{t+1})$, note that
by definition of Bregman divergence, for
some point $\z \in [\x_t, \x_{t+1}]$,
\[  \eqalign{
B_\R(\x_t,\x_{t+1}) =&\ \R(\x_{t}) - \R(\x_{t+1}) - [\nabla\R(\x_{t+1})]^{\top}(\x_t-\x_{t+1})\cr
=&\ \half (\x_t - \x_{t+1})^{\top} \bM_\z (\x_t - \x_{t+1})~.\cr
}\]
By (\ref{eq:9}), assuming $\eta G\Delta< \frac{1}{4}$,
\[
\eqalign{ (\x_t - \x_{t+1})^{\top} \bM_\z (\x_t - \x_{t+1})
  \le&\
   (1- \eta G\Delta)^{-2} (\x_t - \x_{t+1})^{\top} \bM_t (\x_t - \x_{t+1}) \cr
   < &\ 2(\x_t - \x_{t+1})^{\top} \bM_t (\x_t - \x_{t+1}) ~.}
   \]
Thus,
$
B_\R(\x_t,\x_{t+1})
\leq  2 (\x_t - \x_{t+1})^{\top} \bM_t (\x_t - \x_{t+1})
=  2 \eta^2 \g_t^{\top}\bM_t^{-1}\g_t
\leq  2\eta^2\, G^2 \Delta^2 $.
It follows that
\[
\mbox{$R(\x^*)
\leq
  \frac{D^2}{\eta}
+   \frac{1}{\eta} \cdot T \cdot 2\eta^2 G^2 \Delta^2
+ T\cdot  8 G^2 \Delta^2 \cdot \eta
= \frac{D^2}{\eta}
+  10 G^2 \Delta^2  \, T\,\eta $}~.
\]
Thus, picking
$ \eta = \frac{D}{ \sqrt{10} G \Delta}\cdot \frac{1}{\sqrt T} $
yields
$ R (\x^*) \leq
 2 \sqrt{10 } D G \Delta \cdot \sqrt T $.
\end{myproof}

Note that as $\x^*$ tends to the boundary of $\K$, $B_\R(\x^*,\x_1)$ tends to infinity, and hence necessarily so does $D$.
Thus, the regret bound for $\x^*$ on the boundary of $\K$ requires further analysis, as in \cite{AHR08}.
This is what we describe below.

\subsection{The final regret bound}
\begin{Prop} \label{ineq:1}
Let $\R(\cdot)$ be a $\vartheta$-self-concordant barrier for $\K$, and
let $\x\in\interior\K$ and a unit vector $\u\in\reals^n$ be such that $\u^{\top}\nabla\R(\x) > 0$.
Let $t_{\max}= t_{\max}(\x,\u)$ be defined as in {\rm (}\ref{def:tmax}{\rm )}. Under these conditions,
\[ \u^{\top} \nabla\R(\x) \le \frac{\vartheta}{t_{\max}(\x,\u)}~.\]
\end{Prop}
\begin{myproof}
Note that $t_{\max}$ is equal to
the distance from $\x$ to the boundary of $\K$ along the direction of $\u$.
Since $\R(\cdot)$ is a $\vartheta$-self-concordant barrier,
\[  \left[-\frac{1}{\tilde\R'(t)} \right]'
= \frac{\tilde\R''(t)}{[\tilde\R'(t)]^2}
\ge \frac{1}{\vartheta} ~. \]
Consider the function $\tilde\R(t)$ that was defined in (\ref{def:beta}).
For every $t<t_{\max}$,
\[  \frac{1}{\tilde\R'(0)} - \frac{1}{\tilde\R'(t)}
=  \int_0^t \left[\frac{-1}{\tilde\R'(s)} \right]'\,ds \ge \frac{t}{\vartheta} ~\]
and, since $\tilde\beta'(t)$ tends to infinity as $t$ tends to $t_{\max}$, it follows that
$ t_{\max} \le \frac{\vartheta}{\tilde\R'(0)} $.
The latter implies our proposition because $\R'(0) = \u^{\top} \nabla\R(\x)$.
\end{myproof}
For distinct vectors $\x,\y\in \K$, denote
$\tau_{\max}(\x,\y) = t_{\max}\left(\x,\frac{\y-\x}{\|\y-\x\|}\right) $.
\begin{Prop} \label{ineq:2}
If $\x,\y\in\interior\K$ are distinct, then
\[ \R(\y) - \R(\x) \le  - \ln\left( 1 - \frac{\|\y-\x\|}{\tau_{\max}(\x,\y)} \right)\cdot  \vartheta~.\]
\end{Prop}
\begin{myproof}
Given any two distinct points $\x,\y\in\interior \K$, denote $\u=\frac{\y-\x}{\|\y-\x\|}$.
For every $t\in [0, \|\y-\x\|]$,
$ t_{\max}(\x+t\u,\u) = t_{\max}(\x,\u) - t $.
It follows that
\[ \u^{\top}\nabla\R(\x+t\u) \le \frac{\vartheta}{t_{\max}(\x,\u) - t}~. \]
Thus,
\begin{eqnarray*}
~~~~~~\R(\y) - \R(\x)
&=& \int_{0}^{\|\y-\x\|} \tilde\R(t)\,dt
= \int_{0}^{\|\y-\x\|} \u^{\top} \nabla\R (\x+t\u) \,dt \\
&\le & \int_{0}^{\|\y-\x\|} \frac{\vartheta}{t_{\max}(\x,\u) - t} \,dt
 =   -  \ln\left( 1 - \frac{\|\y-\x\|}{\tau_{\max}(\x,\y) } \right)\cdot \vartheta~.
 ~~~~~~~~~~~~~~~~~~
\end{eqnarray*}
\end{myproof}
\begin{Def} [\cite{Nemi}]
Given the initial point $\x_1\in\interior K$ and a real $\delta>0$, the inner subset
$\K(\delta;\x_1)$ is defined by
\[   \K(\delta; \x_1)
= \left\{\y\in\K\, :\,
\|\y-\x_1\| \le \frac{1}{1+\delta}\cdot \tau_{\max}(\x_1,\y)\right\}~.\]
\end{Def}

\begin{Cor}  \label{cor:a}
If $\y\in \K(\delta;\x_1)$, then
$ \R(\y) - \R(\x_1)
\le  \ln\left(1+1/\delta\right)\cdot \vartheta
$.
\end{Cor}
\begin{myproof}
By Proposition \ref{ineq:2},
\begin{eqnarray*}
~~~~~~~~~
\R(\y) - \R(\x_1)
& \le &  -\ln \left( 1 - \frac{\|\y-\x_1\|}{\tau_{\max}(\x_1,\y) } \right)\cdot \vartheta
\le   -\ln \left( 1 - \frac{\tau_{\max}(\x_1,\y)/(1+\delta)}{\tau_{\max}(\x_1,\y) } \right)\cdot \vartheta\\
& = & \ln\left(1+1/\delta\right)\cdot \vartheta~.
~~~~~~~~~~~~~~~~~~~~~~~~~~~~~~~~~~~~~~~~~~~~~~~~~~~~~~~~~~~~~~~~~~~~~~~~~~~~~~~~~
\end{eqnarray*}
\end{myproof}

We are now ready to conclude with the proof of Theorem \ref{thm:main}:
\begin{myproof}[proof of Theorem \ref{thm:main}]
The idea is to analyze the regret vs. $\x^*$ by composing it into regret vs. $\x_\delta^*$ which is in the interior and for which we can use the results of the previous section, as well as the total loss accumulated by taking a close point instead of the exact. This is made precise as follows. 
Given any $\x^*\in\K$ and $\delta>0$, denote
\[ \x^*_\delta = \x_1 + \frac{\tau_{\max}(\x_1,\x^*)}{1+\delta}\cdot \frac{\x^*-\x_1}{\|\x^*-\x_1\|}~.\]
Note that
\[\eqalign{ \|\x^*_\delta - \x^*\|
\le &\tau_{\max}(\x_1,\x^*) -  \frac{\tau_{\max}(\x_1,\x^*)}{1+\delta}
 = \frac{\delta}{1+\delta} \cdot \tau_{\max}(\x_1,\x^*)
 <  \delta\, \Delta~.\cr}\]
We have
\[ \mbox{$
R(\x^*)
= \sum_{t=1}^T f_t(\x_t) - \sum_{t=1}^T f_t(\x^*)
= R(\x^*_\delta)
   + \sum_{t=1}^T\left[ f_t(\x^*_\delta) - f_t(\x^*)\right]$.}
   \]
Also,
\[ \mbox{$
\sum_{t=1}^T\left[ f_t(\x^*_\delta) - f_t(\x^*)\right]
\le \sum_{t=1}^T (\x^*_\delta-\x^*)^{\top}\nabla f_t(\x^*_\delta)
\le \|\x^*_\delta-\x^*\| \cdot \sum_{t=1}^T \|\nabla f_t(\x^*_\delta)\|
 <   \delta\,\Delta \, G\, T$ ~.}
\]
Since $\x^*_{\delta} \in\interior\K$, Theorem \ref{thm:general_norm_past_curvature} can be applied, so
\[ R(\x^*_\delta) \le  8 D G \Delta  \cdot \sqrt T~. \]
Furthermore, since $\x^*_{\delta} \in \K(\delta;\x_1)$, by Corollary \ref{cor:a}
\[\R(\x^*_\delta) - \R(\x_1)
\le  \ln\left(1+1/\delta\right)\cdot \vartheta~ \]
and hence
\[ \eqalign{ B_\R(\x^*_\delta,\x_1)
=&\ \R(\x^*_\delta) - \R(\x_1) - [\nabla^{\top}\R(\x_1)]^{\top} (\x^*_\delta - \x_1)
\leq  \ln\left(1+1/\delta\right)\cdot \vartheta + G\Delta~. \cr
}\]
Thus, may choose $D$ as follows - that will determine $\eta$ in Algorithm \ref{alg:main}, as given in Theorem \ref{thm:general_norm_past_curvature} to be $ \eta = \frac{D}{\sqrt{10} G \Delta }\cdot \frac{1}{\sqrt T}$.
\[ D =   \sqrt{\ln\left(1+1/\delta\right)\cdot \vartheta + G\Delta} \]
so that the regret bound of Theorem \ref{thm:general_norm_past_curvature} with respect to $\x^*_\delta$ holds.
It follows that
\[ R(\x^*) \leq 8 G \Delta \sqrt{\ln\left(1+1/\delta\right)\cdot \vartheta + G\Delta} \cdot \sqrt T
+ \delta\,\Delta \cdot G\, T~.\]
Thus, taking  $\delta = 1/T$, then we obtain 
\begin{eqnarray*}
R(\x^*) & \leq 8 G \Delta \sqrt{\ln\left(1+T \right)\cdot \vartheta  + G\Delta} \cdot \sqrt T + G\Delta \\
& \leq 9  G \Delta \sqrt{\vartheta \ln\left(T \right)   + G\Delta} \cdot \sqrt{T}  & \mbox{ assuming $T > 10 G \Delta$} \\
& \leq 9 \sqrt{\vartheta}  G \Delta \sqrt{T \log T} + 9  (G \Delta)^{3/2}   \sqrt{T}  & \mbox{ $\sqrt{a+b} \leq \sqrt{a} + \sqrt{b} $ for $a,b \geq 0$}
\end{eqnarray*}
\end{myproof}
\paragraph{Remark.}
Since the point $\x_1$ can be chosen arbitrarily, the analysis of the regret over the interval
$1,2,\ldots,T$ can be applied to any subinterval $s,s+1,\ldots,t$. Recall that the algorithm
uses the parameter $ \eta = O(1/\sqrt T)$. It follows that the regret over
any subinterval is also $O(\sqrt{T\log T})$.

\bibliographystyle{plain} 
\bibliography{allrefs} 

\end{document}